\documentclass{segabs}
\usepackage{multirow}
\usepackage{hyperref}       
\usepackage{soul}

\date{}

\usepackage{enumitem}

\usepackage{url}

\usepackage{hyperref} 

\begin{document}

\onecolumn 

\begin{description}[labelindent=-2cm,leftmargin=1cm,style=multiline]

\item[\textbf{Citation}]{C. Zhou, M. Prabhushankar, and G. AlRegib, "Perceptual Quality-based Model Training under Annotator Label Uncertainty," in \textit{International Meeting for Applied Geoscience \& Energy (IMAGE) 2023}, Houston, TX, Aug. 28-Sept. 1, 2023.}

\item[\textbf{DOI}]{\url{https://doi.org/10.1190/image2023-3916384.1}}

\item[\textbf{Review}]{Date of acceptance: May 16, 2023\\Date of presentation: August 29, 2023}

\item[\textbf{Codes}]{\url{https://github.com/olivesgatech/Ramifications-HLU}}

\item[\textbf{Bib}] {@inproceedings\{zhou2023perceptual,\\
  title=\{Perceptual quality-based model training under annotator label uncertainty\},\\
  author=\{Zhou, Chen and Prabhushankar, Mohit and AlRegib, Ghassan\},\\
  booktitle=\{Third International Meeting for Applied Geoscience \& Energy Expanded Abstracts\},\\
  pages=\{965--969\},\\
  year=\{2023\},\\
  doi = \{10.1190/image2023-3916384.1\},\\
  URL = \{https://library.seg.org/doi/abs/10.1190/image2023-3916384.1\},\\
  eprint = \{https://library.seg.org/doi/pdf/10.1190/image2023-3916384.1\},\\
  organization=\{Society of Exploration Geophysicists and the American Association of Petroleum Geologists\}\\
\}
}



\item[\textbf{Keywords}]{Annotator Label Disagreement, Perceptual Quality Assessments, Prediction Uncertainty, Seismic Interpretation} 

\item[\textbf{Contact}]{\href{mailto:chen.zhou@gatech.edu}{chen.zhou@gatech.edu}  OR 
 \href{mailto:mohit.p@gatech.edu}{mohit.p@gatech.edu} OR \href{mailto:alregib@gatech.edu}{alregib@gatech.edu}\\ \url{https://alregib.ece.gatech.edu/} \\ }
\end{description}

\thispagestyle{empty}
\newpage
\clearpage
\setcounter{page}{1}



\title{Perceptual quality-based model training under annotator label uncertainty}

\renewcommand{\thefootnote}{\fnsymbol{footnote}} 

\author{Chen Zhou\footnotemark[1], Mohit Prabhushankar, and Ghassan AlRegib, Center for Energy and Geo Processing (CeGP), School of Electrical and
Computer Engineering, Georgia Institute of Technology,  \{chen.zhou, mohit.p, alregib\}@gatech.edu}

\footer{Example}
\lefthead{Dellinger \& Fomel}
\righthead{Perceptual quality-based model training under annotator label uncertainty}

\maketitle

\begin{abstract}
  Annotators exhibit disagreement during data labeling, which can be termed as annotator label uncertainty. Annotator label uncertainty manifests in variations of labeling quality. Training with a single low-quality annotation per sample induces model reliability degradations. In this work, we first examine the effects of annotator label uncertainty in terms of the model's generalizability and prediction uncertainty. We observe that the model's generalizability and prediction uncertainty degrade with the presence of low-quality noisy labels. Meanwhile, our evaluation of existing uncertainty estimation algorithms indicates their incapability in response to annotator label uncertainty. To mitigate performance degradation, prior methods show that training models with labels collected from multiple independent annotators  can enhance generalizability. However, they require massive annotations. Hence, we introduce a novel  perceptual quality-based model training framework to objectively generate multiple labels for model training to enhance reliability, while avoiding massive annotations. Specifically, we first select a subset of samples with low perceptual quality scores ranked by statistical regularities of visual signals. We then assign de-aggregated labels to each sample in this subset to obtain a training set with multiple labels. Our experiments and analysis demonstrate that training with the proposed framework alleviates the degradation of generalizability and prediction uncertainty caused by annotator label uncertainty.  
  
\end{abstract}

\section{Introduction}
\begin{figure}[t]
\centering
\includegraphics[width=\columnwidth]{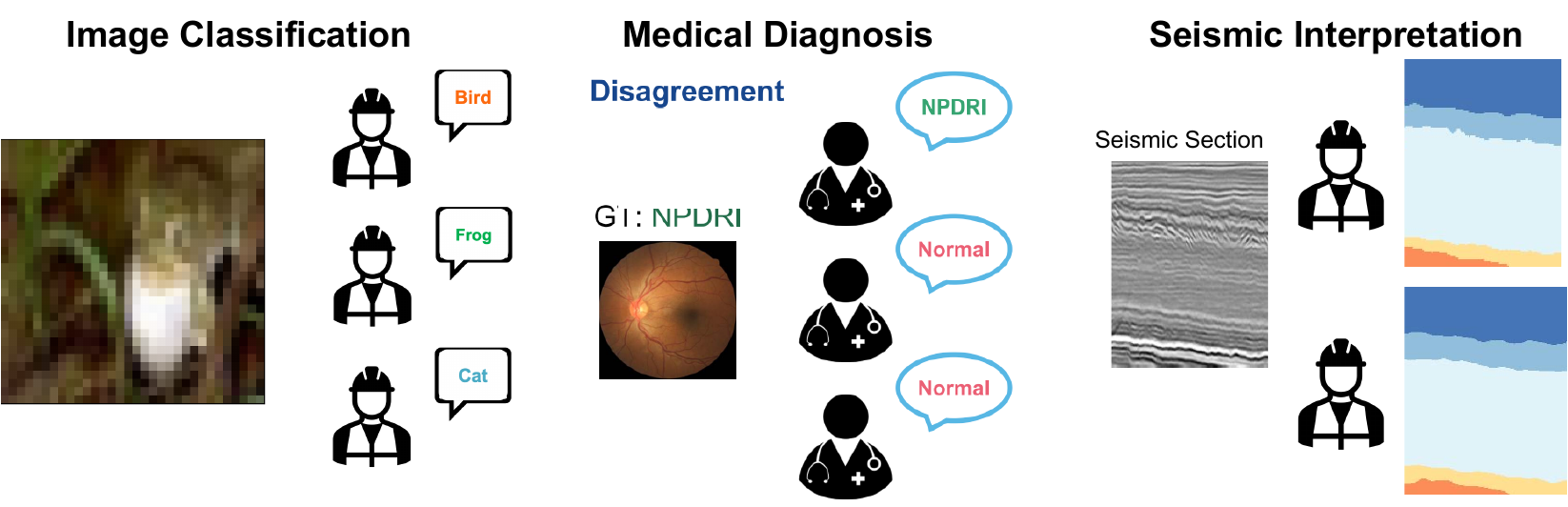}

\caption{The lack of consensus between annotators exists in all data labeling processes across various fields. Left: annotators assign different labels to classify the ambiguous image. Middle: disagreement between doctors in medical diagnosis. Right: interpreters provide different annotations during seismic interpretation.  \vspace{-.3cm}}
\label{fig: annotator_label_uncertainty}
\end{figure}

Raw data usually comes with separate annotations collected from multiple imperfect human annotators. These separate labels lead to a lack of consensus when they disagree with each other. For instance, the authors in \cite{bond2007you} demonstrate different interpretations by multiple geophysicists on the same data, motivating the necessity to study such disagreement and its impact on hydrocarbon predictability.
The lack of consensus between human annotations exists in all data labeling processes and can be critical in various applications, e.g., image classification \cite{wei2021learning}, medical diagnosis \cite{ju2022improving, prabhushankar2022olives}, and seismic interpretation \cite{wang2019residual, benkert2022example}, as shown in Figure ~\ref{fig: annotator_label_uncertainty}. We term this disagreement between human annotators as \emph{annotator label uncertainty}. 

Annotator label uncertainty induces degradation of label quality within machine learning model  development. Training with a single low-quality label per sample leads to model reliability degradation in terms of generalizability and prediction uncertainty. In image classification,  models can overfit to noisy training labels, leading to incorrect class prediction and confidence degradation. In seismic fault prediction, low-quality labels manifest in more diverse modes. There can be false positive labels due to unclear structures. Additionally, substantial faults might be unlabeled due to the difficulty in precisely marking the endpoints of fault picks. These low-quality labels can lead to inaccurate and unconfident predictions by the model.

In order to address the model reliability degradation caused by annotator label uncertainty, the authors in \cite{wei2022aggregate} demonstrate that training with multiple de-aggregated human labels can be beneficial. However, collecting multiple annotations per sample from various independent annotators can be challenging and expensive. To avoid collecting massive subjective manual annotations with disagreement, we aim to generate multiple de-aggregated labels in an objective manner for model training.
The cause of certain data samples that exhibit annotator label uncertainty relates to the difficulty in perception by annotators. In image classification, perceptual quality assessment \cite{moorthy2011blind, temel2016unique, seijdel2020low, prabhushankar2017ms} measures such difficulty. In the geophysics domain, seismic attributes, e.g., gradients of texture (GoT), and seismic saliency, can be utilized to create objective seismic data quality assessments. In this study, we study and tackle annotator label uncertainty in image classification. We propose a perceptual quality-based multi-label training scheme to tackle the model reliability degradation caused by annotator label uncertainty. Specifically, we utilize statistical regularities of visual signals to target the samples that show a high degree of label uncertainty.  For each of these targeted samples, we then assign different labels to perform multi-label training.

We examine the effects of annotator label uncertainty and demonstrate that the model's reliability degradation can be tackled by our proposed perceptual quality-based multi-label training. 
Specifically, we assess the models trained on four label conditions per sample, including a) a single clean label, b) a single noisy label, c) multiple labels collected from different annotators, and d) multiple labels generated objectively by our framework. Compared to the model trained with ideal clean labels, the model trained with noisy labels exhibits higher prediction uncertainty and lower classification accuracy. Our proposed framework avoids model fitting to incorrect sample-label pairs thereby enhancing generalizability and uncertainty. Furthermore, our approach does not require massive manual annotations.

The main contributions of our work include: ($i$) examining the effects of annotator label uncertainty in terms of the model's generalizability and prediction uncertainty. ($ii$) identifying the limitations of certain existing uncertainty estimation algorithms in response to annotator label uncertainty.  ($iii$) introducing a novel perceptual quality-based multi-label training framework to tackle the degradation of generalizability and prediction uncertainty caused by annotator label uncertainty, while not requiring massive human annotations.



\section*{Method}

\begin{figure*}[t]
\centering
\includegraphics[scale=0.5]{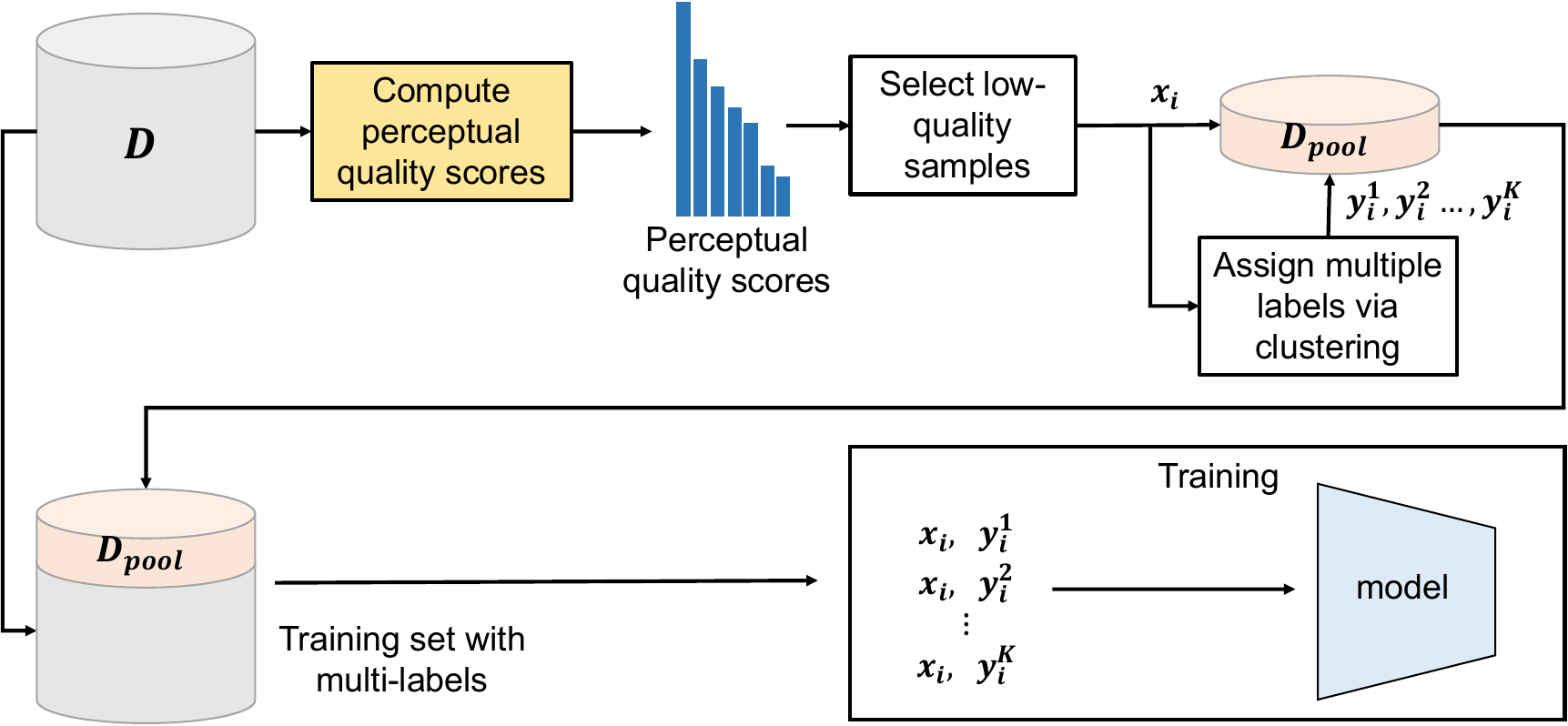}

\caption{Diagram of perceptual quality-based multi-label training framework. We sample a subset $D_{pool}$ consisting of low-quality samples to perform multi-label training in an objective manner. Our framework does not require massive human annotations. $D$ denotes the initial training dataset. $D_{pool}$ denotes the subset targeted by perceptual quality assessments. $x_i$ denotes a sample in $D_{pool}$. $y_i^1, y_i^2,...,y_i^K$ denote multiple labels associated with $x_i$.  \vspace{-.3cm}}
\label{fig: diagram_perceptqlt_train}
\end{figure*}

In this section, we present a novel perceptual quality-based model training scheme that tackles model reliability degradation caused by annotator label uncertainty. At a high level, our framework objectively injects multiple labels into a training subset targeted by perceptual quality assessments.

The overall diagram of our proposed framework is illustrated in Figure ~\ref{fig: diagram_perceptqlt_train}. Consider a model, an initial training dataset $D$. We aim to inject multiple training labels into a subset  $D_{pool}$ that initially contains an underlying single noisy annotation per sample. One can determine $D_{pool}$ by examining the disagreement of annotations collected from multiple independent annotators. However, it is challenging and expensive to collect massive manual annotations. 
The cause of certain samples exhibiting annotator label uncertainty relates to the difficulty of human perception. Objective perceptual quality assessment \cite{moorthy2011blind, temel2016unique, seijdel2020low, prabhushankar2017ms} measures such difficulty. Hence, we propose to exploit perceptual data quality assessments to objectively  determine this subset. We compute the perceptual quality scores of all samples from $D$, which reflect the perception difficulty during labeling. 
Particularly, we employ BRISQUE \cite{mittal2012no} algorithm to obtain perceptual quality scores of all training data. Given an image sample, BRISQUE \cite{mittal2012no} exploits its spatial statistics of locally normalized luminance coefficients to quantify the perceptual quality. The samples with low-quality scores exhibit relatively high annotator label uncertainty and are targeted as $D_{pool}$.
Given that $D_{pool}$ is determined, we  assign multiple de-aggregated labels to mimic annotator disagreement. 
Moderately uncertain samples are likely to be associated with semantically relevant noisy labels, instead of random labels that are completely semantically irrelevant. Hence, we fit a $K$-means model on the training data and assign multiple labels, i.e., $y_i^1, ..., y_i^k, k\in \{1,2,..., K\}$ to each sample $x_i \in D_{pool}$ based on its feature distance to the clustering centroids. $K$ denotes the cardinality of the set of multiple labels. The remaining samples are associated with clean labels. We combine $D_{pool}$ with the remaining clean data to train the model. Note that for a sample $x_i \in D_{pool}$, the model is given its $k$ replications, each associated with one of the labels drawn from $y_i^1, ..., y_i^k$.  









\section{Experiments}

We first examine the effects of noisy annotations on the model's reliability in terms of generalizability and prediction uncertainty. We then show the efficacy of the proposed framework in enhancing the model reliability under annotator label uncertainty. 

In order to examine the effects of noisy annotations, we compare the model's generalizability and prediction uncertainty between two training label conditions, a) training with a single clean label per sample, and b) training with a single noisy human label per sample. Specifically, we use the original CIFAR-10 \cite{krizhevsky2009learning} training images for both training label conditions, and use the noisy human annotations in CIFAR-10-N \cite{wei2021learning} for noisy label condition b). 
Following the uncertainty evaluation protocols in prior works \cite{gal2016dropout, hendrycks2019augmix, zhou2021amortized, prabhushankar2022introspective}, we assess predictive uncertainty and generalizability of a vanilla ResNet18 (Vanilla), Monte Carlo (MC)  Dropout \cite{gal2016dropout}, and deterministic uncertainty quantification (DUQ) \cite{van2020uncertainty} under two types of distribution shifts, including rotation and corruption. For all models, we use the same base ResNet-18 \cite{he2016deep}. We measure generalizability and prediction uncertainty via accuracy and entropy \cite{shannon1948mathematical}, respectively. 
We summarize the prediction uncertainty and generalizability of these models on the test sets of CIFAR-10-Rotations and CIFAR-10-C \cite{hendrycks2019benchmarking} in the first two rows (Clean and Human Noise) of both the top and bottom parts in Table ~\ref{Tab:uncertain_acc_hlu}. 
We observe significant degradation in both prediction uncertainty and accuracy by comparing the performance of the models trained with clean labels, and noisy human annotations. Besides, we find that MC Dropout and DUQ models are not able to tackle annotator label uncertainty since they do not enhance the performance of uncertainty measures compared to the vanilla network.

We then verify the efficacy of the proposed framework in tackling the uncertainty and generalizability degradation. The core of the framework is to construct the uncertain training subset $D_{pool}$ with multiple labels. We ensure that the number of samples in $D_{pool}$ is the same as the number of samples with noisy labels in CIFAR-10-N. Specifically, we rank all training samples in $D$ via perceptual quality statistics and select the top $40\%$ of uncertain samples to construct $D_{pool}$. Furthermore, we ensure that the level of label uncertainty in the constructed $D_{pool}$ is the same as that of human labeling. All samples in $D_{pool}$ are associated with maximum $K=3$ labels. For each sample in $D_{pool}$ with the top $10\%$ uncertainty measured by perceptual quality scores, we assign at least two noisy labels. The remaining $30\%$ of samples are assigned with one noisy label. In this way, we construct $D_{pool}$ to have the same overall label disagreement as that of single noisy training label condition  with the real-world noisy annotations in CIFAR-10-N.
 
Overall, the degradation in prediction uncertainty under the noisy label condition can be mitigated by our perceptual quality-based multi-label training framework as shown in the last row of both the top and bottom parts in Table ~\ref{Tab:uncertain_acc_hlu}. In addition to mitigating the increase in predictive uncertainty, training with our framework also enhances generalizability as measured in classification accuracy. These observations validate that our proposed framework mitigates the negative effects caused by annotator label uncertainty. We also include the results of an annotator-based multi-label training scheme using real-world manual annotations collected from multiple independent annotators \cite{wei2021learning} as shown in the third row in Table ~\ref{Tab:uncertain_acc_hlu}. While training with multiple manual labels collected from different annotators achieves the highest accuracy, it requires massive human annotations and does not show better enhancement in prediction uncertainty. This observation confirms the efficacy of utilizing perceptual quality-based multi-label training to enhance the model’s prediction uncertainty.


\begin{table*}[t]
\centering
\caption{Evaluation results on CIFAR-10-Rotations (top part) and CIFAR-10-C (bottom part). The three uncertainty quantification models are assessed against different training label conditions. Overall, our proposed perceptual quality-based multi-label training framework mitigates the degradation of generalizability and prediction uncertainty caused by annotator label uncertainty, while avoiding massive manual annotations. All models are trained with the original CIFAR-10 training images but with different labels.}
\begin{tabular}{ccc|cc|cc}
\hline
\multirow{4}{*}{Training Label Conditions} & \multicolumn{6}{c}{CIFAR-10-Rotations}                               \\ \cline{2-7} 
                                 & \multicolumn{2}{c}{Vanilla} & \multicolumn{2}{c}{MC Dropout} & \multicolumn{2}{c}{DUQ} \\
                                 &        \begin{tabular}{@{}c@{}}Entropy \\ $(\downarrow)$\end{tabular}      & \begin{tabular}{@{}c@{}}Accuracy \\ $(\uparrow)$\end{tabular}      &        \begin{tabular}{@{}c@{}}Entropy \\ $(\downarrow)$\end{tabular}      & \begin{tabular}{@{}c@{}}Accuracy \\ $(\uparrow)$\end{tabular}        &        \begin{tabular}{@{}c@{}}Entropy \\ $(\downarrow)$\end{tabular}      & \begin{tabular}{@{}c@{}}Accuracy \\ $(\uparrow)$\end{tabular}    \\ \hline 
 &  \multicolumn{6}{c}{\textbf{Effects of Annotator Label Uncertainty}} \\   \hline
Clean             &  0.675       & 0.446            & 0.696    & 0.446         &   1.071     & 0.427   \\  
Human Noise                 &      2.398       & 0.355            &  2.413      & 0.350          & 1.697    & 0.282    \\  \hline 
&  \multicolumn{6}{c}{\textbf{Multi-label Training Frameworks}} \\  \hline

Human-based                    & 2.069                & 0.392                        & 2.058                        & 0.394                      & 2.068                       & 0.379        \\



\textbf{Perceptual quality-based (ours)}                   & {1.264}                & {0.362}                         & {1.284}                       & {0.357}                  & {1.569}                       & {0.350}       \\
\hline 

\hline
\multirow{4}{*}{Training Label Conditions} & \multicolumn{6}{c}{CIFAR-10-C}                               \\ \cline{2-7} 
                                 & \multicolumn{2}{c}{Vanilla} & \multicolumn{2}{c}{MC Dropout} & \multicolumn{2}{c}{DUQ} \\
                                 &        \begin{tabular}{@{}c@{}}Entropy \\ $(\downarrow)$\end{tabular}      & \begin{tabular}{@{}c@{}}Accuracy \\ $(\uparrow)$\end{tabular}      &        \begin{tabular}{@{}c@{}}Entropy \\ $(\downarrow)$\end{tabular}      & \begin{tabular}{@{}c@{}}Accuracy \\ $(\uparrow)$\end{tabular}        &        \begin{tabular}{@{}c@{}}Entropy \\ $(\downarrow)$\end{tabular}      & \begin{tabular}{@{}c@{}}Accuracy \\ $(\uparrow)$\end{tabular}    \\ \hline  
&  \multicolumn{6}{c}{\textbf{Effects of Annotator Label Uncertainty}} \\   \hline
Clean           & 0.384    & 0.750            & 0.396     & 0.746         &  0.829    & 0.686   \\  
Human Noise                 &  2.172   & 0.640           &  2.184     & 0.634         &   1.563      & 0.470       \\  \hline 
&  \multicolumn{6}{c}{\textbf{Multi-label Training Frameworks}} \\  \hline
Human-based                     & 1.855               & 0.702                         & 1.848                       & 0.706                      & 1.863                        & 0.664       \\   


\textbf{Perceptual quality-based (ours)}                   & {1.222}                 & 0.580                         & {1.243}                       & 0.580                   & {1.496}                       & 0.540       \\
\hline 
\end{tabular}
\label{Tab:uncertain_acc_hlu}
\end{table*}

\section{Conclusions}
In this work, we first examine the effects of annotator label uncertainty on the model's generalizability and prediction uncertainty. Meanwhile, we observe that certain existing uncertainty quantification algorithms are incapable of responding to annotator label uncertainty. 
We introduce the concept of exploiting perceptual quality assessments, which  objectively quantify annotation quality, within a multi-label training framework to enhance generalizability and prediction uncertainty under annotator label uncertainty. The experiments on an image classification dataset validate that our proposed training framework can improve classification accuracy and prediction entropy of all three uncertainty quantification models with the presence of noisy labels while avoiding the collection of massive manual annotations. For seismic-related applications, we suggest that the concept of perceptual quality-based multi-label training can be applied to the geophysics domain by exploiting seismic attributes. Specifically, one can utilize seismic attributes, e.g., grey-level co-occurrence matrix (GLCM) texture, gradients of texture (GoT), and seismic saliency, to create objective seismic data quality assessments. The seismic quality assessments can be utilized to enhance the model's reliability under annotator label uncertainty without requiring massive annotations.

\section{ACKNOWLEDGMENTS}

This work is supported by the ML4Seismic Industry Partners at Georgia Tech.










\twocolumn

\bibliographystyle{seg}  
\bibliography{example}

\end{document}